\documentclass{article}
\usepackage{ijcai25}

\usepackage{times}
\usepackage{soul}
\usepackage{url}
\usepackage[hidelinks]{hyperref}
\usepackage[utf8]{inputenc}
\usepackage[small]{caption}
\usepackage{graphicx}
\usepackage{amsmath}
\usepackage{amsthm}
\usepackage{booktabs}
\usepackage{algorithm}
\usepackage{algorithmic}
\usepackage[switch]{lineno}
\title{Building Patient Journeys in Hebrew:\\ A 
Language Model for Clinical Timeline Extraction}

\author{
Kai Golan Hashiloni$^1$
\and
Brenda Kasabe Nokai$^2$\and
Michal Shevach$^2$\and
Esthy Shemesh$^2$\and
Ronit Bartin$^2$\and
Anna Bergrin$^2$\and
Liran Harel$^2$\and
Nachum Dershowitz$^3$\and
Liat Nadai Arad$^2$\and
Kfir Bar$^1$\\
\affiliations
$^1$Efi Arazi School of Computer Science, Reichman University, Herzilya, Israel\\
$^2$Tel Aviv Sourasky Medical Center, Israel\\
$^3$School of Computer Science and AI, Tel Aviv University, Israel\\
\emails
kai.golanhashiloni@post.runi.ac.il,
\{brendak, michalshevach, esthyshe, roni, annaber, liranha, liatna\}@tlvmc.gov.il, nachum@tau.ac.il, kfir.bar@runi.ac.il
}
\begin{document}

\maketitle
\begin{abstract}
We present a new Hebrew medical language model designed to extract structured clinical timelines from electronic health records, enabling the construction of patient journeys. Our model is based on DictaBERT 2.0 and continually pre-trained on over five million de-identified hospital records. To evaluate its effectiveness, we introduce two new datasets—one from internal medicine and emergency departments, and another from oncology—annotated for event temporal relations. Our results show that our model achieves strong performance on both datasets. We also find that vocabulary adaptation improves token efficiency and that de-identification does not compromise downstream performance, supporting privacy-conscious model development. The model is made available for research use under ethical restrictions.
\end{abstract}

\section{Introduction}

Electronic health records (EHRs) provide a longitudinal account of a patient’s clinical journey—capturing symptoms, diagnoses, treatments, and outcomes over time. Much of this information is stored in unstructured clinical notes, where the temporal ordering of medical events is essential for understanding disease progression, evaluating treatment efficacy, and supporting clinical decision-making. However, modeling this temporal structure remains a significant challenge, particularly in languages and settings with limited resources.

In this work, we tackle the task of structuring patient journeys by formulating it as temporal relation classification (TRC) over EHRs. TRC involves identifying the temporal relationships between medical events mentioned in free-text clinical notes—such as determining whether one event occurred before, after, or at the same time as another. For example, from the sentence ``The patient developed chest pain before receiving aspirin'', a TRC system should infer that the event \textit{chest pain} occurred before the event \textit{aspirin administration}. Accurate TRC enables the construction of coherent patient timelines, which in turn support downstream applications such as clinical summarization, disease progression modeling, and diagnosis prediction. It also serves as a foundation for temporal reasoning in clinical decision support systems, helping to contextualize and prioritize interventions based on the sequence of events.

While recent advances in large language models (LLMs) have led to substantial gains on general NLP tasks, these models often underperform on TRC. In particular, Roccabruna \textit{et al.}~\shortcite{roccabruna-etal-2024-will} demonstrated that supervised encoder-based models outperform generative LLMs on this task. Moreover, several studies have shown that adapting LLMs to specific domains improves their performance on related downstream tasks \cite{MedBERT,chen2023meditron70bscalingmedicalpretraining,gema-etal-2024-parameter}.

Our goal in this work is to build a practical, real-world TRC model tailored to the medical domain and trained on authentic patient records. In our medical center all clinical documentation is written in Hebrew, therefore we focus specifically on the Hebrew language—a low-resource setting that has been largely neglected in previous work in this area. 
Existing medical language models have been trained almost exclusively on English data, leaving languages like Hebrew with little to no support. To address this gap, we introduce the first Hebrew medical TRC model trained on real clinical notes, and we evaluate it on two newly constructed TRC tasks that span different areas of medical practice and reflect real patient care trajectories.

Our training procedure involves two main steps. First, we develop HeMed, an encoder-based language model pre-trained on a corpus of 5 million de-identified clinical notes from a large hospital. We initialize with a general-purpose Hebrew language model and continue pre-training on this clinical corpus, exploring techniques such as vocabulary adaptation and improved word segmentation. Second, we fine-tune and evaluate HeMed on two novel Hebrew TRC datasets, which we manually annotate for this purpose.

This work not only addresses the scarcity of Hebrew medical NLP resources, but also highlights the importance of temporal modeling in EHRs for capturing and reasoning about patient trajectories.

To summarize, the main contributions of this work are:
\begin{itemize}
\item We introduce the first medical temporal relation classification (TRC) model for Hebrew, trained and evaluated on real-world clinical notes from the Tel Aviv Sourasky Medical Center (TSMC, notioned as \textit{}{the hospital)}. This model supports the temporal structuring of patient journeys and enables time-aware clinical NLP applications in a low-resource setting.

\item We compile the first large-scale corpus of clinical notes in Hebrew (5 million de-identified records), and use it to adapt a foundational Hebrew language model through continued pre-training. The resulting model, \textbf{HeMed}, is made available to the research community for medical NLP in Hebrew.

\item We design and annotate two novel Hebrew TRC datasets reflecting different clinical domains, providing the first benchmark for evaluating temporal reasoning capabilities in Hebrew EHRs.

\item We show that continued pre-training on clinical notes provides significant performance improvements over the original language model.

\end{itemize}

\section{Related Work}
\label{ssec:related_work}

\subsection{Temporal Relationship Classification}
While temporal relation classification (TRC) has received substantial attention in the English-language NLP literature, Hebrew and other low-resource languages remain significantly underexplored \cite{Hebrew-TRC}. One of the most prominent datasets for English TRC is MATRES \cite{ning-etal-2018-multi}, which includes 12,736 training instances and 837 test instances. 
% Other notable English TRC datasets include TempEval \cite{verhagen-etal-2007-semeval}, TimeBankDense \cite{chambers-etal-2014-dense}, RED \cite{ogorman-etal-2016-richer}, and TCR \cite{ning-etal-2018-joint}.

A wide range of computational approaches have been proposed for TRC in English. Many rely on language models \cite{han2021econeteffectivecontinualpretraining,han-etal-2019-deep,han-etal-2019-joint,roccabruna-etal-2024-will,eirew2025pairwiseglobalzeroshottemporal}, while others incorporate global inference mechanisms to ensure temporal consistency \cite{ning-etal-2019-improved,mathur-etal-2021-timers}, including applications in the medical domain \cite{zhou2020clinicaltemporalrelationextraction}. Several methods also leverage linguistic features extracted from text \cite{zhang-etal-2022-extracting,wang-etal-2023-extracting}, and graph-based approaches have gained traction—for instance, models that encode syntactic and semantic relations using graph neural networks \cite{zhou-etal-2022-rsgt} or syntax-guided graph transformers \cite{zhang-etal-2022-extracting}. 

In contrast, Hebrew TRC research is still in its infancy. To date, only one publicly available TRC dataset exists for Hebrew \cite{Hebrew-TRC}, containing 7,260 annotated samples.

% \subsection{Medical LLMs} \label{ssec:medical_llms}
% Several medical LLMs have been publicly released in recent years.
% Early encoder models like MedBERT \cite{MedBERT} and PubMedBERT \cite{BLUERB-PubmedBERT} were trained from scratch on scientific medical literature.
% Other models, such as BioBERT \cite{BioBERT}, SciBERT \cite{Scibert}, and BlueBERT \cite{BlueBERT}, were developed through continual pre-training of a general-purpose language model on domain-specific data.
% Non-English medical encoder models, such as the Swedish clinical BERT \cite{clinical-kbbert} and MedGBERT for German \cite{MedGBERT}, have followed similar continual pre-training strategies.

% Recent advances include larger architectures like MedPaLM2 \cite{MedPaLM2}, MedGemini \cite{MedGemini}, and Medical mT5 \cite{medicalmt5opensourcemultilingual}, demonstrating the efficacy of multilingual and multimodal fine-tuning. 

% Many medical models have been trained using the LLama model family. 
% One such model is Meditron \cite{chen2023meditron}, which was trained on a large corpus of scientific literature in English. 
% More recently, a new family of models, Me-LLama \cite{MedLlama}, was developed based on LLama-2, incorporating both scientific literature and clinical notes from the MIMIC \cite{johnson2023mimic} dataset.

% All these models are typically evaluated on well-established medical downstream tasks, such as MedQA \cite{jin2021disease} and MIMIC-CXR \cite{johnson2019mimic}. 
% Notably, all these medical benchmarks are available only in English.

Recent findings show that encoder-based models continue to outperform other architectures on the TRC task \cite{roccabruna-etal-2024-will}. These models also tend to be smaller and more efficient, making them well-suited for real-time TRC systems intended for deployment in clinical settings.
To apply TRC for timeline extraction, one can use a pairwise approach, in which the model predicts temporal relations for each pair of events individually; the resulting labeled pairs are then combined into a temporal graph representing the event timeline. This method is commonly referred to as \textit{pairwise} TRC. In contrast, \textit{global} TRC aims to predict the full temporal graph of a document in a single step. However, encoder-based models have been shown to perform poorly on global TRC, and since our study focuses on encoder architectures, we restrict our approach to pairwise TRC.

\subsection{Domain Adaptation of LLMs}

We begin building our model by continually pre-training an existing Hebrew base model on clinical notes. During this phase, we explore various strategies to enhance the model's adaptation to the clinical domain.

Generic language models often underperform in specialized domains, highlighting the importance of domain adaptation \cite{domainadaptationgeneralizationpretrained}. A widely used approach is continual pre-training, where a pre-trained model is further trained on domain-specific data. This strategy underlies many successful models, including the popular SciBERT \cite{Scibert} and LEGAL-BERT \cite{legalBERT} models.

Tokenizer adaptation is another critical component, as standard tokenizers may poorly segment domain-specific terms. Techniques such as exBERT \cite{tai-etal-2020-exbert} and vocabulary expansion in Chinese LLaMA \cite{ChineseLlama,llama} improve token efficiency by adding domain-relevant tokens. AdaLM \cite{adalm} extends this by iteratively adding tokens until corpus-level coverage stabilizes, improving both representation and compression. Related methods include AVocaDo \cite{Avocado} and vocabulary replacement strategies \cite{csaki2023efficientlyadaptingpretrainedlanguage}. For Hebrew, DictaLM-2.0 \cite{DictaBERT2.0} adapted Mistral-7B by adding 1,000 Hebrew tokens, reducing the compression rate by half.

In this work, we systematically explore two vocabulary adaptation methods and two segmentation approaches across two clinical tasks.

\section{Methods}
\label{sec:approach}

Our approach consists of two main steps: 1) We perform continual pre-training on an existing encoder-based foundation model using a newly compiled corpus of Hebrew clinical notes extracted from EMRs; 2) We fine-tune the resulting medical language model on a temporally annotated TRC dataset.

The pre-training objective in the first step mirrors that of the original foundation model—masked language modeling (MLM), where the model learns to predict masked tokens within a sequence. To further enhance performance, we experiment with several vocabulary adaptation strategies and evaluate their impact across the two downstream tasks we use in Step 2. 

Several Hebrew models can serve as the Hebrew foundation language model, including HeBERT \cite{hebert}, AlephBERT (AB) \cite{seker-etal-2022-alephbert}, AlephBERTGimmel (ABG) \cite{abg}, and DictaBERT 2.0 \cite{DictaBERT2.0}. 
Preliminary experiments showed that HeBERT performed worse than the others, while DictaBERT 2.0 slightly outperformed ABG. 
ABG, as an extension of AB, has been shown to outperform AB across various tasks.
Therefore, we use DictaBERT 2.0 as the base model for all subsequent experiments.

Encoder-based models typically have fewer parameters, making them easier to train by requiring fewer optimization steps and less training data.
Their main advantage is that they can be efficiently fine-tuned for specific tasks using internal training equipment typically available in hospitals.
% BERT-based models feature an encoder-only architecture, which prevents them from performing generative tasks. 
% Instead, they are primarily designed for classification tasks at the token or sequence level.
% There are many practical medical tasks that such a model can perform. 
% For example, it can extract lists of medications and diseases from a medical record or generate a timeline of events from textual EMRs to reconstruct a patient’s journey.
% In the next section, we outline the process of collecting the corpus used for training the model.

\subsection{Step 1: Continual Pre-training on Medical Records} \label{ssec:corpus}
\subsubsection{Collecting Data}
\begin{figure}[t]  %[htbp]
    \centering
    \includegraphics[width=2.5in]{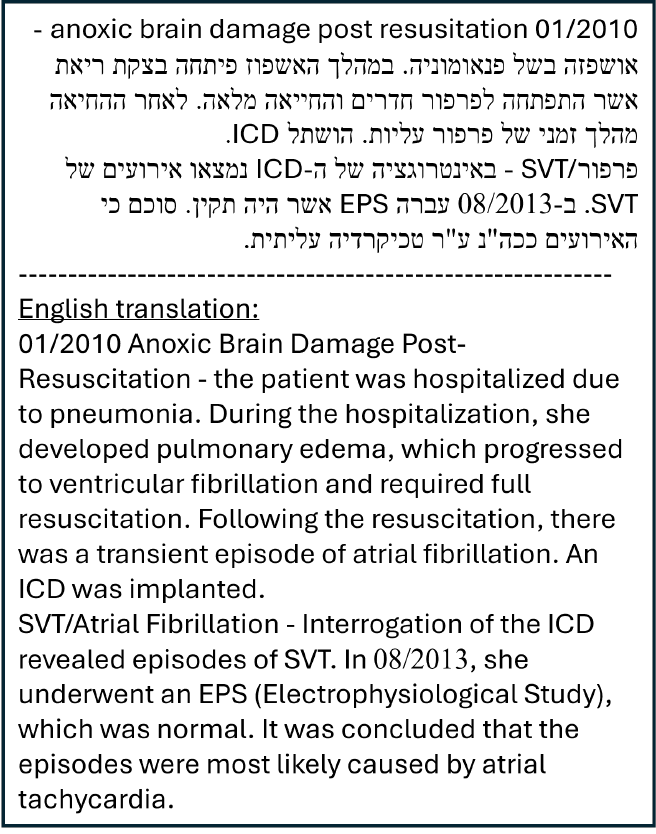}
  \caption{Example of an EMR from our corpus along with its English translation.}
  \label{fig:record}
\end{figure}

To obtain high-quality real-world data for training the model, we collected textual EMRs from the hospital.
There are several key considerations when collecting data from a hospital’s internal system.
First, we aim for diversity in both topics and medical practices to train a general medical language model. 
Ensuring a broad representation of medical contexts helps improve the model's adaptability across different clinical scenarios.
Second, we seek to minimize duplication as much as possible. A common practice in internal clinics, where patients often return for multiple visits, is for caregivers to copy the previous visit’s record and update it based on the patient’s current status.
This practice significantly increases duplication, reducing the effective diversity of the training data.  
Third, we aim to include different types of records with varying lengths. 
To address these considerations, we collected EMRs from multiple hospital divisions, including internal medicine departments, the ER, the children's ER, and the ER for labor and delivery. 
These divisions exhibit lower duplication rates than other departments and clinics while offering a high degree of diversity in topics and medical practices.  
The dataset includes various types of textual EMRs, such as visit summaries, status descriptions, discharge letters, follow-ups, and diagnoses. 
This variety ensures a broad range of text lengths and vocabulary distributions.  
Overall, our corpus consists of approximately five million records (3GB of uncompressed data), containing around 0.5 billion words, which correspond to approximately 0.7 billion tokens when tokenized using DictaBERT’s tokenizer.
Detailed corpus statistics are provided in Table~\ref{tab:corpus}.
Figure~\ref{fig:record} shows an example of a single record, provided with its English translation.
We present some of the most common words in the corpus in Supplementary B.

Since EMRs may contain sensitive private information, we opted to de-identify them by automatically removing identifying details and replacing them with generic names and terms while preserving the natural flow of the text. 
Overall, we replaced 2,335,950 substrings during the de-identification process.
Additionally, we assess the impact of this de-identification process on the model’s performance, which we discuss further in the next section.

\begin{table}[tp]
  \centering
  {\caption{Corpus statistics: Word counts are estimated by splitting all strings based on white spaces. A complete breakdown by department is provided in Supplementary F.} \label{tab:corpus}}
  {\begin{tabular}{lr}
  \toprule
  \bfseries Info & \bfseries Count \\
  \midrule
  Records             & 5,232,028       \\
  Words  & 546,606,802           \\
  Characters       & 3,185,476,400           \\
  \bottomrule
  \end{tabular}}
\end{table}

% \subsection{Model Training}
% Our training focuses on three key aspects: vocabulary adaptation, segmentation, and de-identification. 
% We experimented with different combinations of approaches and compared their performance.

\subsubsection{Vocabulary Adaptation} \label{sssec:da}
We explored several methods to adapt the model and its vocabulary to the medical records.  
Each tokenizer was trained on one million randomly sampled records from our pre-training corpus.

The first approach, referred to as the \textbf{simple} method, follows the strategy used in the ChineseLLama model \cite{ChineseLlama}.
In this method, we trained a new tokenizer with a vocabulary of 10,000 tokens on our domain-specific corpus, using the same tokenizer configuration as DictaBERT, which is based on the WordPiece \cite{wu2016googlesneuralmachinetranslation} algorithm.
We then expanded DictaBERT’s tokenizer vocabulary by performing a set union with these 10,000 newly learned tokens. 
DictaBERT’s original vocabulary consists of 128K tokens, so this modification introduces a relatively small expansion. 
Additionally, since some of the 10K tokens may already exist in DictaBERT’s vocabulary, the actual number of newly added tokens is lower. 
We deliberately kept the number of added tokens limited, as prior research has shown that significantly increasing vocabulary size does not necessarily improve performance \cite{tai-etal-2020-exbert}.
We present some of the most frequent tokens added to the tokenizer in Supplementary C.

The second approach follows the \textbf{AdaLM} method proposed by \cite{adalm}, which we previously discussed in Section~\ref{sssec:da}. 
In this case, we used \(\delta=0.1\) with a step size of 1,000 tokens.

We initialized the embeddings of the new tokens using the average of the embeddings of their sub-tokens, as described in \cite{nag-etal-2023-entropy-embedding-init}. 
The authors of that study reported no significant difference in performance among several initialization methods but found that random initialization led to suboptimal results. 

\subsubsection{Segmentation} \label{sssec:semgentation}
Segmentation is the process of converting input text into a sequence of tokens based on a given vocabulary. 
Given a text, the tokenizer typically has multiple ways to segment it. 
The most widely used segmentation technique, commonly paired with WordPiece, selects the longest possible token from the beginning of a word. 
Words are defined during the pre-tokenization step (usually based on white spaces).
In addition to the standard WordPiece segmentation, we explored FLOTA \cite{flota}, which greedily selects the longest available token in a word, regardless of its position. 

\begin{figure}[t]
    \centering
    \includegraphics[width=3in]{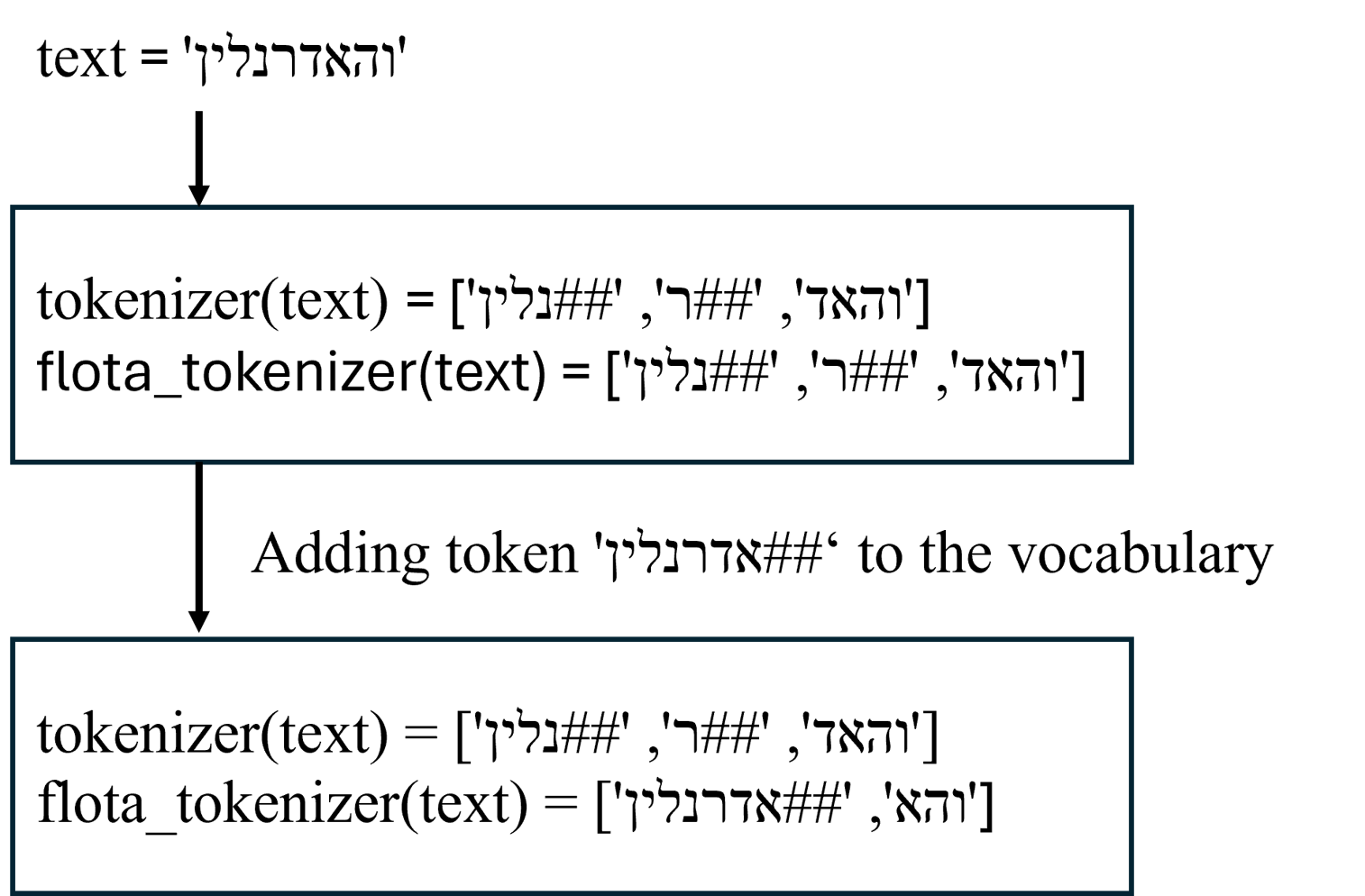}
  \caption{An example of text segmentation with and without FLOTA, before and after domain adaptation. The text translates to ``and the adrenaline''. In Hebrew, ``and'' and ``the'' are prefixed to the word as particles.}
  \label{fig:flota_example}
\end{figure}

In the original paper, FLOTA was designed for fine-tuning and inference time, independent of the segmentation method used during pre-training.
However, in this work, we apply it also to the continual pre-training step, to make sure the relevant tokens get a better domain-specific representation.
To illustrate the need for FLOTA, we show an example in Figure~\ref{fig:flota_example}. FLOTA ensures that tokens introduced through vocabulary adaptation are actively used, allowing their embeddings to be trained and refined, rather than remaining unused.

\subsubsection{De-identification}
\label{sssec:deidentification}
Since our ultimate goal is to publicly release both the new medical pre-trained language model and the medical TRC model, it was essential to ensure that all records used for training were properly de-identified.
We began by using an open-source de-identification system for Hebrew medical texts, called Safe Harbor.\footnote{\url{https://github.com/8400TheHealthNetwork/HebSafeHarbor}} This system is based on a named entity recognition (NER) model, enhanced with a rule-based engine. 
The NER model is trained to detect and classify entities such as names, locations, and dates.

However, manual evaluations revealed two key issues. 
First, the system sometimes missed certain names and locations, leaving sensitive information exposed. 
Second, it exhibited over-masking, replacing non-private spans with placeholders unnecessarily. 
This over-masking altered the text more than required, disrupting natural language patterns and potentially degrading model performance. 
For example, the system misclassified lab results (e.g., ``120--90'' for blood pressure) as identity numbers and mistakenly replaced disease names, procedures, and drug names with personal name placeholders.

To address these issues, we developed a custom rule-based de-identification procedure that focuses only on the essential information that must be concealed. 
Our approach leverages metadata available for each EMR, allowing more precise de-identification of entities such as names, phone numbers, identity numbers, dates, and addresses.
Specifically, we retrieved relevant patient information from the hospital’s internal databases, such as the names of the patient, their relatives, and listed contacts. 
We then search for these names in the medical record using regular-expression rules, accounting for Hebrew’s rich morphology. 
This process is fully automated.

Similarly, we extracted doctor names, identifying numbers, email addresses, and zip codes, ensuring they are accurately detected. 
Organizational names are masked only if they refer to medical institutions or other known organizations. 
To achieve this, we compiled a manual list of institutions to look for.

Instead of inserting generic placeholders (e.g., $<$phone$>$), we implemented smart replacements to preserve the text’s natural structure. 
When possible, detected entities are replaced with random but realistic alternatives (e.g., another fake phone number instead of $<$phone$>$). 
This approach reduces information loss, maintains syntactic fluency, and improves the model’s ability to learn from natural-looking data.

\subsection{Step 2: Fine-tuning on TRC} \label{sssec:datasets}
We annotated two new datasets for the downstream task of TRC, following the same annotation guidelines as the MATRES dataset \cite{ning-etal-2018-multi}, which was later adapted to Hebrew in \cite{Hebrew-TRC}.
We see TRC as a key task in medical applications. 
By leveraging TRC labels, one can construct a timeline of medical events within a textual record, effectively summarizing a patient’s journey. 
This structured representation can support predicting medical recommendations, enhance clinical decision-making, and facilitate visual summarization of complex medical histories.

In this task, events are marked in the text in advance, and each anchored pair of events \((e_1, e_2)\) is assigned one of the following labels: \textbf{before}: \( e_1 \) occurred before \( e_2 \) in chronological order;
\textbf{after}: \( e_1 \) occurred after \( e_2 \);  
\textbf{equal}: \( e_1 \) and \( e_2 \) occurred simultaneously; and, 
\textbf{vague}: the temporal relationship between \( e_1 \) and \( e_2 \) cannot be determined.  

We created two datasets of EMRs annotated for TRC. 
The first is \textbf{Med-TRC}, derived from the same EMR corpus used for continual pre-training. 
We ensured that the records selected for TRC annotation were excluded from the pre-training process.
The second dataset, \textbf{Onc-TRC}, was collected from a completely different medical domain—visit reports from the hospital’s breast oncology clinics. 
This dataset allows us to evaluate the model’s performance on a domain distinct from the one used during pre-training.

Two professional annotators, both experienced nurses with a medical background, labeled the two datasets. 
In cases of uncertainty, they consulted a team of medical experts.

To assess annotation quality, 15\% of the records were annotated by both annotators independently. 
Cohen’s Kappa agreement score for Onc-TRC was 0.8.  
For Med-TRC, while we did not formally measured Kappa, qualitative assessment indicated a high level of agreement between annotators.

Before annotation, we marked the events in the texts.
For Med-TRC, we followed the approach described in \cite{Hebrew-TRC}, where events were identified as verbs using a Hebrew part-of-speech (POS) tagger from the Stanza library.\footnote{\url{https://stanfordnlp.github.io/stanza}}
For Onc-TRC, we use a predefined list of oncological terms that, according to the medical experts we consulted, represent the most important events in an oncology patient’s record. 
This list includes procedures, medications, biomarkers, and other key medical terms, covering about 300 terms.
To identify these terms in the textual records, we utilized a simple matching algorithm that detects all occurrences of the listed terms. 
Additionally, we used regular expressions to handle basic morphological variations in Hebrew, ensuring better coverage and accuracy.
Following \cite{Hebrew-TRC}, only event pairs within a sliding window of two sentences (determined by hard punctuation) were annotated. 
For each event pair, presented with its surrounding text as context, annotators were asked to assign one of the four temporal labels described earlier.
Since both the POS tagger used for event identification in Med-TRC and the simple matching algorithm used in Onc-TRC can make mistakes, we added an additional label, \textbf{invalid}, for cases where at least one event was incorrectly identified by the respective algorithm.

After filtering out invalid pairs, we ended up with 4,262 valid pairs for Med-TRC and 2,377 for Onc-TRC.
The distribution of labels across both datasets is shown in Table~\ref{tab:trc-distribution}.
There is a significantly higher number of \textit{before} pairs. 
This is expected given the chronological nature of language and aligns with findings from the MATRES dataset, where similar distributions have been observed.

\begin{table}[tp]
\centering
\small
  {\caption{TRC datasets label distribution.}\label{tab:trc-distribution}}
  {\begin{tabular}{lrr}
  \toprule
  \bfseries Label & \bfseries Med-TRC & \bfseries Onc-TRC\\
  \midrule
  BEFORE     & 2,756    & 1,432    \\
  AFTER      & 826      & 381    \\
  EQUAL      & 572      & 416     \\
  VAGUE      & 108      & 148      \\
  \bottomrule
  \end{tabular}}
\end{table}

Due to the relatively large number of \textit{before} instances, the task is somewhat easy, allowing even the base models to perform well. 
To mitigate this disproportion when evaluating different language model variations, we clipped the TRC datasets, reducing the number of \textit{before} instances to match the count of the second most common class.
After clipping and splitting into train and test sets: Med-TRC contains 1,829 training samples and 503 test samples, and Onc-TRC contains 1,078 training samples and 283 test samples.

To fine-tune each language model variation on the TRC task, we followed the approach proposed in \cite{Hebrew-TRC}, specifically using the event start state (ESS) architecture, which was reported to perform slightly better than alternative architectures.
For evaluation, we used the weighted average F1 metric. 
As in the original study, we also adopted the relaxed F1 score, which disregards errors where non-\textit{vague} predictions are made on \textit{vague} instances.

Unfortunately, these two datasets cannot be released to the public, as they may contain sensitive information.

\section{Results on Real Data} 

In Step 1, we began by training and evaluating several tokenizer variants, following the vocabulary adaptation strategies outlined in Section~\ref{sssec:da}.
 
As mentioned earlier, we use 1 million records for training each tokenizer. 
For the validation set, we set aside 10,000 records from the original corpus to analyze tokenizer performance.
We initialized all tokenizers using the DictaBERT tokenizer configuration \cite{DictaBERT2.0}, but with an empty vocabulary. 
After training, we merged the newly learned vocabulary with DictaBERT’s original vocabulary.  
To assess tokenizer performance, we sampled the 1-million-records training corpus five times using different random seeds and calculated the average of two key metrics: corpus token count (CTC) \cite{schmidt2024tokenizationcompression}, which has been reported to correlate with downstream performance, as well as compression rate, as proposed in \cite{goldman2024unpackingtokenizationevaluatingtext}.  
Detailed explanations of these metrics can be found in Supplementary D.  

As mentioned in Section~\ref{sssec:da}, the simple adaptation approach merges a 10K-token vocabulary into DictaBERT’s existing vocabulary, though some tokens were already present. 
The actual number of new tokens added in this approach is 2,011. 
In the AdaLM approach, 3,000 new tokens were added.

As shown in Table~\ref{tab:tokenizers-res}, the vocabulary adaptation methods generally reduce the number of tokens required to represent medical text, resulting in more efficient compression by the tokenizers.
Additionally, we observe that AdaLM improves both metrics compared to the simple adaptation approach. 
We recognize that CTC and compression rate are not necessarily strong predictors of downstream task performance, as noted in \cite{schmidt2024tokenizationcompression}.
Therefore, in the next section, we directly evaluate the different model variations on the three downstream tasks.

\begin{table}[tp]
\small
\centering
  {\caption{CTC and compression rate (CR) for different vocabulary adaptation methods on the 10K test split of our corpus. The first row represents the original DictaBERT vocabulary, serving as a baseline.}\label{tab:tokenizers-res}}
  {\begin{tabular}{lrr}
  \toprule
  \bfseries Tokenizer & \bfseries CTC & \bfseries CR\\
  \midrule
  DictaBERT & 1,339,002 & 1.45    \\
  \midrule
  Simple  & 1,318,682  & 1.43     \\
  Simple+FLOTA  & 1,321,248 & 1.43 \\
  AdaLM  & 1,232,328  & \textbf{1.34}     \\
  \bottomrule
  \end{tabular}}
\end{table}

\begin{figure*}[t]
    \centering
    \includegraphics[width=0.9\linewidth]{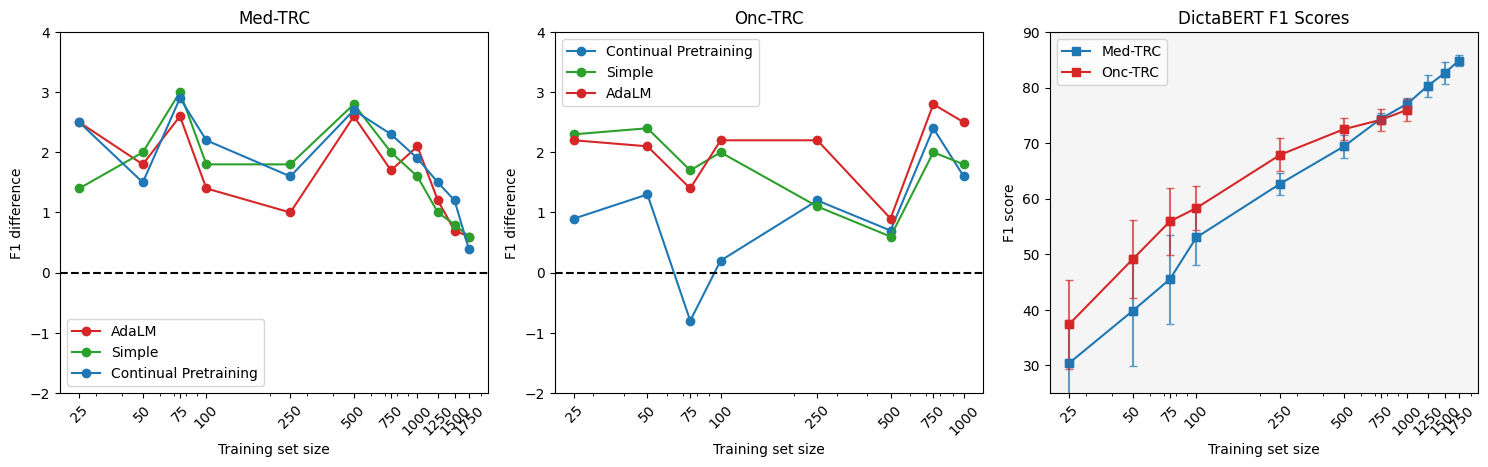}
    \caption{F1-score difference (in absolute points) from the baseline model across various training set sizes (in number of instances). All models were pre-trained on the full corpus before fine-tuning on the downstream tasks. The rightmost figure shows the absolute mean F1-score and the standard deviation of the \textbf{baseline} model.}
  \label{fig:final}
\end{figure*}

\begin{figure*}[t]
    \centering
    \includegraphics[width=0.9\linewidth]{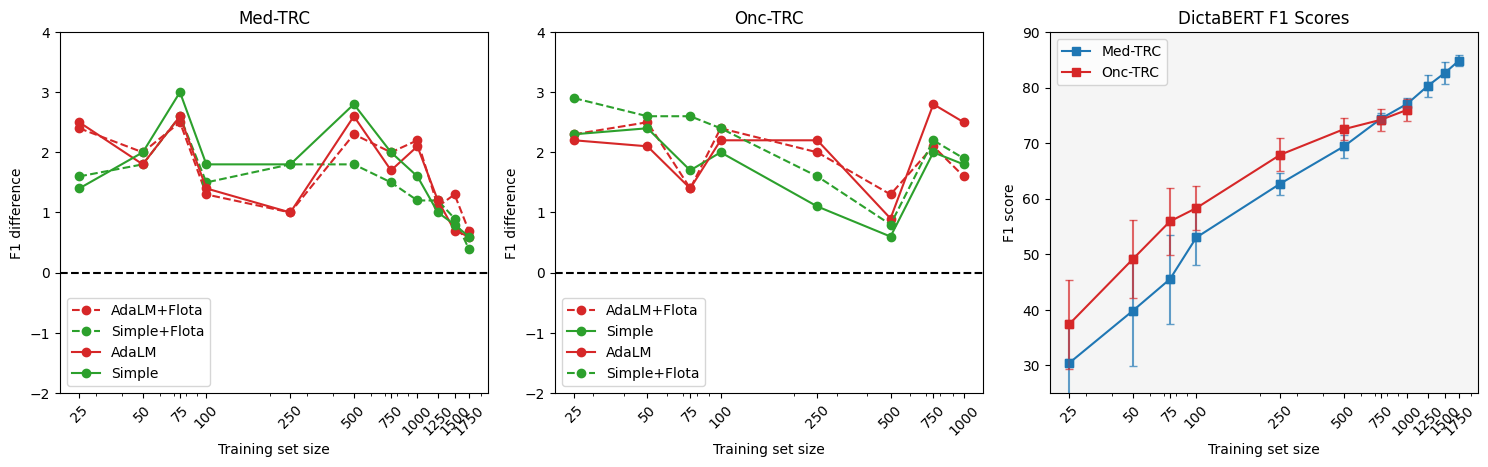}
    \caption{F1-score difference (in absolute points) from the baseline model across various training dataset sizes (in number of instances). Simple adaptation with and without FLOTA. The rightmost figure shows the absolute mean F1-score and the standard deviation of the \textbf{baseline} model.}
    \label{fig:flota}
\end{figure*}

\begin{figure*}[t]
    \centering
    \includegraphics[width=0.9\linewidth]{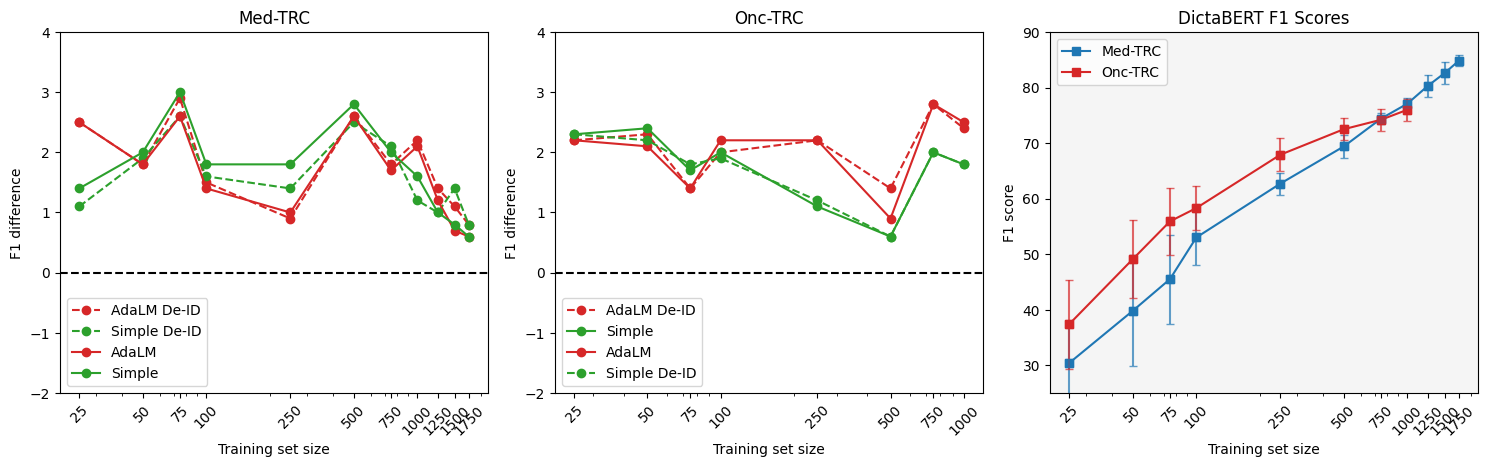}
  \caption{F1-score difference per training dataset size. Simple vocabulary-adaptation model pre-trained with and without de-identification. The rightmost figure shows the absolute mean F1-score and the standard deviation of the \textbf{baseline} model.}
  \label{fig:anon}
\end{figure*}

\subsection{TRC Evaluation}
We evaluate the models' performance on the two downstream TRC tasks using varying training-set sizes. 
Specifically, we repeat the experiments with training sizes of 25, 50, 75, 100, 250, and so on.
This approach allows us to measure the impact of the continual pre-training on the amount of labeled data required for effective downstream medical task performance.
The importance of this step is not only for a deeper comparison but also to show how many samples are needed to attain a certain degree of improvement over the baselines.
In the medical field, where data is scarce and private, annotation is often complex and expensive. 
Therefore, a model that can learn effectively from fewer examples is highly valuable. 

To ensure robustness, we repeated each fine-tuning process 15 times with different random seeds and report the mean performance across all runs. 
The complete results are presented in Supplementary A, including standard deviations, which are all within the range of 1-2 minor points. 
Additionally, we ran multiple English medical encoder models on the two downstream tasks as general baselines. 
All English models underperformed compared to our Hebrew baseline models.
For convenience, we report only the difference in the relevant performance metric relative to the base model (DictaBERT 2.0), which was fine-tuned on each task but did not undergo the continual pre-training procedure described in Step 1.
In other words, for each experiment, we run DictaBERT 2.0 alongside all adapted versions and calculate the difference in performance. 
% Each experiment is repeated 15 times with different seeds, and the average performance is computed. 
% The reported difference is calculated between the averaged results across all runs.

All models were trained on the full clinical-notes corpus using a learning rate of \(3\times 10^{-6}\), which we empirically found to be optimal. 
Training was conducted with a batch size of 16 samples on a single NVIDIA A40 GPU (48GB of GPU RAM), with each epoch taking approximately 63 hours.

The entire code and the supplementary material are released alongside this paper under an Apache-2.0 license and can be found in the project's repository.\footnote{\url{https://github.com/golankai/HeMed-TRC}}

The results for the three model variations—standard continual pre-training (without vocabulary adaptation), simple vocabulary adaptation, and AdaLM—are visualized in Figure~\ref{fig:final}.
In both tasks, the newly trained models consistently outperform the baseline model.  
In the Med-TRC task, the improvement diminishes as the training set size increases, suggesting that the benefits of continual pre-training are more pronounced in low-data scenarios.  
However, in the Onc-TRC, the performance gain remains consistent, at least for some of the variations, even as the training set size grows, indicating that the adapted models provide a robust advantage across different data availability conditions.
We observe that no single model variation consistently outperforms the others across the two datasets.  
In Med-TRC, the models achieve similar performance, with AdaLM lagging slightly behind the other variations.  
However, in Onc-TRC, the trend shifts, and AdaLM outperforms the other methods by a small margin.

\subsection{Segmentation}
We trained our model using the simple vocabulary adaptation method, once with the default WordPiece segmentation and once with the FLOTA segmentation algorithm, applying it during both continual pre-training and fine-tuning. 
As shown in Figure~\ref{fig:flota}, using FLOTA does not significantly impact the model’s performance across tasks.

\subsection{Number of Epochs}
All reported results are based on one epoch of continual pre-training. 
We extended this to two and three epochs to assess the impact of longer training. 
Additional epochs slightly improve Med-TRC but have no effect on Onc-TRC.
The results are provided in Supplementary E.

\subsection{De-identification}
We wanted to assess the impact of pre-training on the de-identified corpus, as described in Section~\ref{sssec:deidentification}.
In Figure~\ref{fig:anon}, we compared the performance of the simple vocabulary-adaptation model trained on both the de-identified corpus and the original corpus. 
The results show that despite replacing approximately 2.3M substrings, accounting for about 5\% of the words in the corpus,  performance degradation is minimal.
This suggests that de-identification does not significantly impact the model’s effectiveness.

\section{Discussion} 
\label{sec:conclusions}
In this work, we introduced a Hebrew medical language model based on DictaBERT 2.0, with the primary goal of advancing clinical timeline extraction from EHRs. This capability is essential for structuring patient journeys, a foundational step in supporting longitudinal clinical analyses and decision-making.

To evaluate the model’s effectiveness, we constructed two novel datasets: one derived from internal medicine and ER records, and another from oncology. These datasets enabled us to assess the model’s performance across diverse clinical settings. Our results show that the proposed model performs well on both datasets, demonstrating its ability to capture temporal relations and support robust patient timeline construction.
The performance we achieved is comparable to results reported on standard English pairwise TRC datasets.

Beyond the task-specific contributions, our study also investigated core design decisions in building domain-specific language models. We found that vocabulary adaptation reduces tokenization overhead, enabling more efficient use of the model's fixed 512-token context window and allowing it to process longer and more meaningful clinical segments. This efficiency gain is particularly valuable for EHR narratives, which often include densely packed and time-sensitive information.
Additionally, continual pre-training on de-identified domain-specific data significantly improved performance across clinical tasks. These findings echo trends observed in other languages and domains, further validating continual adaptation as a key component in clinical language-modeling development. Crucially, our experiments show that de-identification, as applied in our pipeline, does not degrade model performance—highlighting a viable path toward privacy-conscious medical model development without compromising utility.
To support ethical research, our model is released with usage restrictions: 
It is available for research purposes only, and may be used solely by formal medical institutions under IRB-approved protocols.

\appendix

\section*{Ethical Statement}

This research was approved by the Helsinki Ethical Review Board (IRB) of Tel Aviv Souraski Medical Center (0668-23 TLV).

\section*{Acknowledgments}
This research was partially funded by Grant No.\@ 81670 from the Israel Innovation Authority.
We would like to thank Tamar Almogy for her assistance during our work.

% The preparation of these instructions and the \LaTeX{} and Bib\TeX{}
% files that implement them was supported by Schlumberger Palo Alto
% Research, AT\&T Bell Laboratories, and Morgan Kaufmann Publishers.
% Preparation of the Microsoft Word file was supported by IJCAI.  An
% early version of this document was created by Shirley Jowell and Peter
% F. Patel-Schneider.  It was subsequently modified by Jennifer
% Ballentine, Thomas Dean, Bernhard Nebel, Daniel Pagenstecher,
% Kurt Steinkraus, Toby Walsh, Carles Sierra, Marc Pujol-Gonzalez,
% Francisco Cruz-Mencia and Edith Elkind.

\bibliographystyle{named}
\bibliography{ijcai25}

\end{document}